%% file: main.tex
\documentclass[10pt,twocolumn,letterpaper]{article}
\usepackage[accsupp]{axessibility}
\usepackage{cvpr}              
\usepackage{graphicx}
\usepackage{amsmath}
\usepackage{amssymb}
\usepackage{booktabs}
\usepackage{multirow}
\usepackage{xcolor}
\usepackage{colortbl}

\definecolor{mypink}{rgb}{.99,.91,.95}
\usepackage{pifont}
\usepackage[pagebackref,breaklinks,colorlinks]{hyperref}
\usepackage{xcolor}         
\newcommand{\cmark}{\ding{51}\xspace}%
\newcommand{\xmark}{\ding{55}\xspace}%

\usepackage[multiple]{footmisc}
\usepackage{bigfoot}
\DeclareNewFootnote{AAffil}[arabic]
\DeclareNewFootnote{ANote}[fnsymbol]
\usepackage[title]{appendix}

\usepackage[capitalize]{cleveref}
\crefname{section}{Sec.}{Secs.}
\Crefname{section}{Section}{Sections}
\Crefname{table}{Table}{Tables}
\crefname{table}{Tab.}{Tabs.}

\def\cvprPaperID{6161} 
\def\confName{CVPR}
\def\confYear{2023}

\begin{document}

\title{CapDet: Unifying Dense Captioning and Open-World Detection Pretraining}

\author{Yanxin Long$^{1}$\footnotemark[1] \quad Youpeng Wen$^{1}$\footnotemark[1]  \quad Jianhua Han$^{2}$ \quad Hang Xu$^{2}$ \quad Pengzhen Ren$^{1}$\\ Wei Zhang$^{2}$ \quad Shen Zhao$^{1}$\footnotemark[2] \quad Xiaodan Liang$^{1,3}$\footnotemark[2]\\
{\normalsize $^{1}$Shenzhen Campus of Sun Yat-sen University \quad $^{2}$Huawei Noah's Ark Lab \quad $^{3}$MBZUAI}\\
{\tt\small longyx9@mail2.sysu.edu.cn, wenyoupeng0@outlook.com, hanjianhua4@huawei.com, }\\
{\tt\small chromexbjxh@gmail.com, renpzh@mail.sysu.edu.cn, wz.zhang@huawei.com, }\\
{\tt\small z-s-06@163.com, xdliang328@gmail.com}}

\maketitle

\renewcommand{\thefootnote}{\fnsymbol{footnote}} 
\footnotetext[1]{Equal contribution.} 
\footnotetext[2]{Corresponding authors.} 

\input{paper_files/0-abstract}
\input{paper_files/1-introduction}
\input{paper_files/2-related_work}

\input{paper_files/3-methods}

\input{paper_files/4-experiment}
\input{paper_files/5-conclusion}
\begin{appendices}
\input{paper_files/appendix.tex}

\end{appendices}

\clearpage

{\small
\bibliographystyle{ieee_fullname}
\bibliography{egbib}
}

\end{document}

%% file: paper_files/0-abstract.tex
\vspace{-10pt}
\begin{abstract}
\vspace{-5pt}

Benefiting from large-scale vision-language pre-training on image-text pairs, open-world detection methods have shown superior generalization ability under the zero-shot or few-shot detection settings. However, a pre-defined category space is still required during the inference stage of existing methods and only the objects belonging to that space will be predicted. To introduce a “real” open-world detector, in this paper, we propose a novel method named CapDet to either predict under a given category list or directly generate the category of predicted bounding boxes. Specifically, we unify the open-world detection and dense caption tasks into a single yet effective framework by introducing an additional dense captioning head to generate the region-grounded captions. Besides, adding the captioning task will in turn benefit the generalization of detection performance since the captioning dataset covers more concepts. Experiment results show that by unifying the dense caption task, our CapDet has obtained significant performance improvements (e.g., +2.1\% mAP on LVIS rare classes) over the baseline method on LVIS (1203 classes). Besides, our CapDet also achieves state-of-the-art performance on dense captioning tasks, e.g., 15.44\% mAP on VG V1.2 and 13.98\% on the VG-COCO dataset.

\end{abstract}
\vspace{-4pt}

%% file: paper_files/1-introduction.tex
\vspace{-5mm}
\section{Introduction}
\label{sec:intro}

Most state-of-the-art object detection methods \cite{redmon2016you, ren2015faster, zhu2020deformable} benefit from a large number of densely annotated detection datasets (\textit{e.g.}, COCO \cite{lin2014microsoft}, Object365 \cite{shao2019objects365}, LVIS \cite{gu2021open}). However, this closed-world setting results in the model only being able to predict categories that appear in the training set. Considering the ubiquity of new concepts in real-world scenes, it is very challenging to locate and identify these new visual concepts. This predictive ability of new concepts in open-world scenarios has very important research value in real-world applications such as object search \cite{meng2015object, philbin2007object}, instance registration \cite{zhang2017deep}, and human-object interaction modeling \cite{gkioxari2018detecting}.

\begin{figure}[t!]
		\begin{center}

\includegraphics[width=1\linewidth]{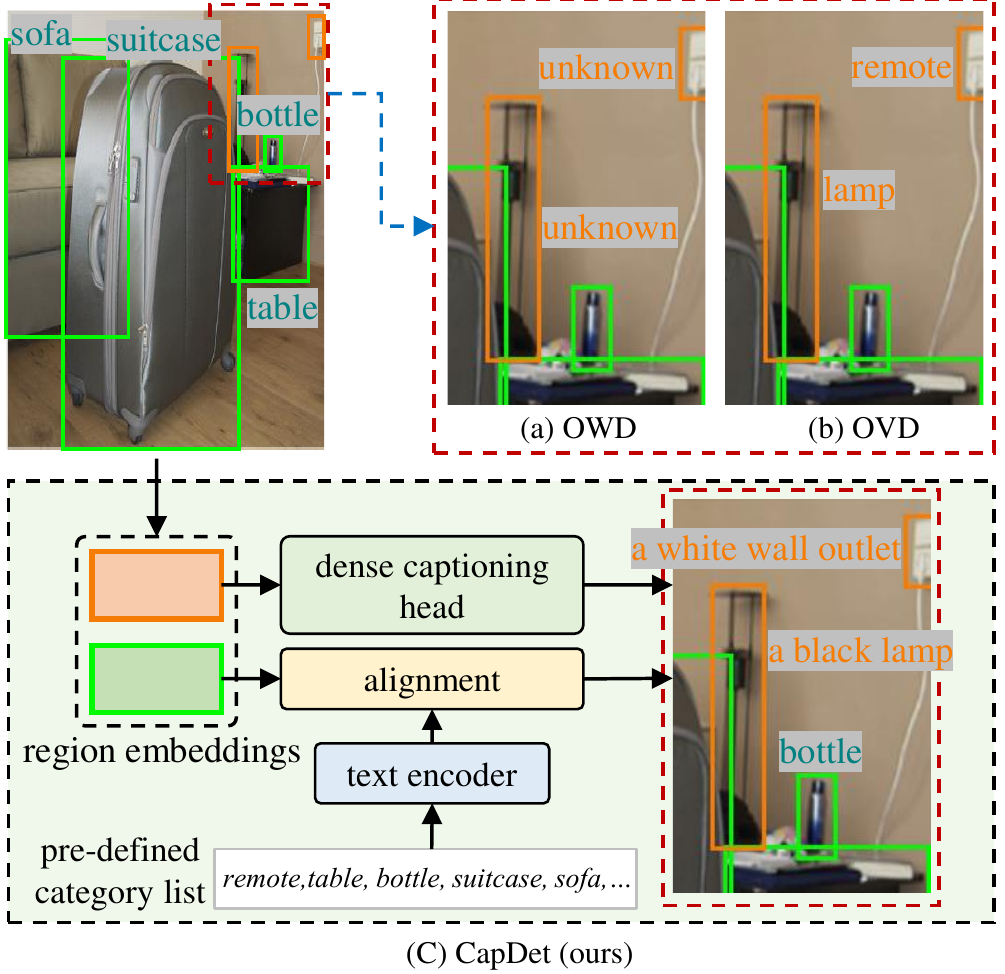}
		\end{center}
		\vspace{-6mm}
		\caption{Comparison of the different model predictions under OWD, OVD, and our setting. (a) OWD methods~\cite{joseph2021towards, gupta2022ow, zhao2022revisiting} are not able to describe the detailed category of the detected unknown objects and (b) the performance of OVD methods~\cite{gu2021open, du2022learning,yao2022detclip} usually depends on the pre-defined category list during the inference. (c) With the unification of two pipelines of dense captioning and open-world detection pre-training, 
		our CapDet can either predict under a given category list or directly generate the description of predicted bounding boxes.}
		\vspace{-8mm}
		\label{fig:intro_figure}
\end{figure}


Currently, the open world scenario mainly includes two tasks: \textit{open world object detection}~\cite{joseph2021towards} (OWD) and \textit{open-vocabulary object detection}~\cite{zareian2021open} (OVD).
Although the paradigms of OWD and OVD tasks are closer to the real world, the former cannot describe the specific concept of the detected unknown objects and requires a pre-defined category list during the inference. 
Specifically, as shown in Figure~\ref{fig:intro_figure}, 
previous OWD methods~\cite{joseph2021towards, gupta2022ow, zhao2022revisiting} would recognize new concepts not in the predefined category space as “unknown". Further, another line of task OVD requires the model to learn a limited base class and generalize to novel classes. Compared to the \textit{zero-shot object detection} (ZSD) proposed by~\cite{rahman2018zero}, OVD allows the model to use external knowledge, \textit{e.g.}, knowledge distillation from a large-scale vision-language pre-trained model~\cite{gu2021open, du2022learning}, image-caption pairs~\cite{zareian2021open}, image classification data~\cite{zhou2022detecting}, grounding data~\cite{yao2022detclip, zhang2022glipv2, li2022grounded}. With the external knowledge, OVD methods show a superior generalization capacity to detect the novel classes within a given category space.
However, as shown in Figure \ref{fig:intro_figure}, when given an incomplete category list, OVD can only predict the concepts that appear in the given category list, otherwise, there will be recognition errors, ( \textit{i.e.}, as illustrated in Figure \ref{fig:intro_figure} (b), the OVD methods prone to predict the  “wall socket” as “remote”, since the latter is in the category list but not the former).

Thus, under the OVD setting, we mainly face the following two challenges: (\textit{i}) it is difficult to define a complete list of categories; (\textit{ii}) low response values on rare categories often lead to recognition errors.
This is mainly because we cannot exhaustively enumerate new objects in the real world, and secondly, it is difficult to collect enough samples for rare classes.
However, the fact that rare objects in the real world, even some new objects that are unknown to humans, such as UFOs, do not prevent people from using natural language to describe it as “a flying vehicle that looks like a Frisbee".

Therefore, based on the above observations, in this paper, we consider a new setting that is closer to the open world and real scenes, \textit{i.e.}, we expect the model to both detect and recognize concepts in a given category list, and to generate corresponding natural language descriptions for new concepts or rare categories of objects.
Early dense captioning methods \cite{johnson2016densecap, gao2022caponimage} can locate salient regions in images and generate the region-grounded captions with natural language.
Inspired by this,  to address the challenges faced in the OVD setting, we propose to unify the two pipelines of dense captioning and open-world detection pre-training into one training framework, called \textbf{CapDet}.
It empowers the model with the ability to both accurately detect and recognize common object categories and generate dense captions for unknown and rare categories by unifying the two training tasks.



Specifically, our CapDet constructs a unified data format for the dense captioning data and detection data. With the data unification, CapDet further adopts a unified pre-training paradigm including open-world object detection  and dense captioning pre-training. 
For open-world detection pretraining, we treat the detection task as a semantic alignment task and adopt a dual encoder structure as~\cite{yao2022detclip} to locate and predict the given concepts list. 
The concepts list contains category names in detection data and region-grounded captions in dense captioning data. 
For dense captioning pretraining, CapDet proposes a dense captioning head to take the predicted proposals as input to generate the region-grounded captions. 
Due to the rich visual concepts in the dense captioning data
, the integration of dense captioning tasks will in turn benefit the generalization of detection performance. 

Our experiments show that the integration of few dense captioning data brings in large improvement in the object detection datasets LVIS, 
\textit{e.g.}, +2.7\% mAP on LVIS. The unification of dense captioning and detection pre-training gains an additional 2.3\% increment on LVIS and 2.1\% increment on LVIS rare classes. Besides, our model also achieves state-of-the-art performance on dense captioning tasks. 
Note that our method is the first to unify dense captioning and open-world detection pretraining.

To summarize, our contributions are three folds:
\begin{itemize}
    \vspace{-1mm}
    \item We propose a novel open-vocabulary object detection framework CapDet, which cannot only detect and recognize concepts in a given category list but also generate corresponding natural language descriptions for new concept objects.
    \vspace{-2mm}
    \item We propose to unify the two pipelines of dense captioning and open-world detection pre-training into one training framework. Both two pre-training tasks are beneficial to each other.
    \vspace{-2mm}
    \item Experiments show that by unified dense captioning task and detection task, our CapDet gains significant performance improvements on the open-vocabulary object detection task (\textit{e.g.}, +3.3\% mAP on LVIS rare classes). Furthermore, our CapDet also achieves state-of-the-art performance on the dense captioning tasks, \textit{e.g.}, 15.44\% mAP on Visual Genome (VG) V1.2 and 13.98\% mAP on VG-COCO.
\end{itemize}
\vspace{-6mm}



%% file: paper_files/2-related_work.tex
\section{Related Work}
\paragraph{Vision-Language Pre-training.}
Vision-Language Pre-training \cite{radford2021learning_clip, jia2021scaling_align, dong2022maskclip} as a scheme in the domains of natural language processing \cite{bert,gpt3} and computer vision \cite{vit} obtains continual attention currently. 
And it exhibits strong performance and generalization ability on various downstream vision and cross-modal tasks.
Among them, CLIP \cite{radford2021learning_clip} and ALIGN \cite{jia2021scaling_align} as dual-stream methods utilize large-scale image-text pairs on the Internet by cross-modal contrastive learning to get excellent zero-shot classification ability. Single-stream methods \cite{kim2021vilt,li2019visualbert} unify visual and textual embeddings in a single transformer-based model, which can perform text generation tasks such as image caption and VQA. Some mixed architectures \cite{wang2021vlmo,li2022blip} combine single-stream and dual-stream to explore a unified way of vision-language understanding and generation. However, these methods take low-resolution images as input and serve the task of classification and retrieval. Those vision-language pre-training approaches can not be applied to pure computer vision task directly, \textit{i.e.}, object detection task. 
\vspace{-4mm}
\paragraph{Open World Object Detection / Open-Vocabulary Object Detection.}
Object detection is a core computer vision task, which aims at localizing objects using a bounding box and classifying them. The mature detection approaches which show great performance on supervised data include one-stage detectors (\textit{i.e.}, YOLO \cite{redmon2016you}, ATSS \cite{zhang2020bridging}) having a relatively high detection efficiency and two-stage detectors (\textit{i.e.}, Faster R-CNN \cite{ren2015faster}, Mask R-CNN \cite{he2017mask}) having good detection accuracy. However, how to generalize these methods to rare classes and novel concepts in the real world is a challenge. Currently, several object detection approaches for such open-world scenes have attracted extensive attention from academia and industry. These methods are divided into two tasks which are called open-world object detection and open-vocabulary object detection respectively depending on whether to detect the class of unknown classes. 

For the OWD task, Zhao et al.\cite{zhao2022revisiting} proposed a proposal advisor to 
assist in identifying unknown proposals without supervision and a class-specific expelling classifier to filter out confusing predictions. 
For the OVD task, GLIP\cite{li2022grounded} converts the detection data into grounding format and proposes a fusion module to learn semantic vision information in grounding data. K-Lite\cite{shen2022k} reconstructs the input format of the data in GLIP from sequential to parallel and uses nouns hierarchy and definition to format text sequence. DetCLIP\cite{yao2022detclip} unifies detection, grounding, and image-text pair data in a paralleled formulation and constructs a concept dictionary to augment the text data, which strikes a balance between performance and efficiency. Differing from all these works, our CapDet can generate an open-set caption of each region proposal to cover situations where the semantics of new object instances are not in the given category list. 
\vspace{-2mm}
\paragraph{Dense Captioning.}
Dense captioning aims at generating detailed descriptions for local regions, which usually needs to locate visual regions with semantic information and generate captions for these regions. J. Johnson et al.\cite{johnson2016densecap} utilized a fully convolutional localization network to locate regions of interest (RoIs) and then describe them. Afterward, many methods\cite{yin2019context,li2019learning} based on Faster-RCNN\cite{ren2015faster} and LSTM\cite{graves2012long} are proposed to do dense captioning. X. Li et al.\cite{li2019learning} arrange RoI features as a sequence and put them into LSTM with the guidance of the region features to form the complementary object context features. This method also needs ground truth bounding boxes auxiliary tests to achieve good results. But limited by the forget gate mechanism of LSTM, the inputted sequence cannot be too long. Then, the transformer-based method TDC\cite{shao2022region} is proposed to tackle the long sequence forgotten problem. 
Instead, our CapDet proposes a transformer-based caption head to generate a caption using a single-stage detector ATSS while simultaneously achieving open-world detection.

%% file: paper_files/3-methods.tex
\begin{figure*}[t!]
		\begin{center}
            \includegraphics[ width=0.85\linewidth]{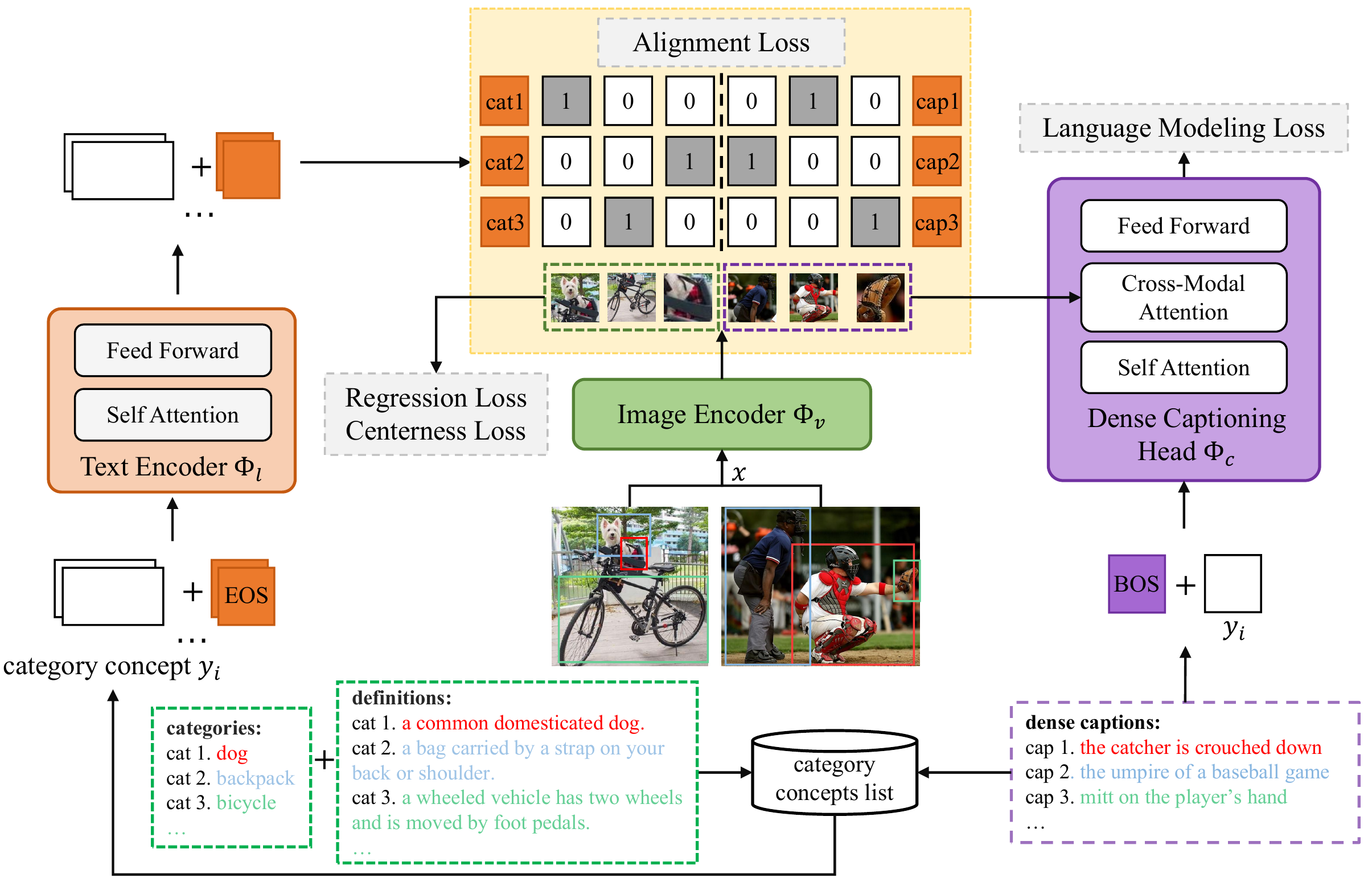}
		\end{center}
		\vspace{-4mm}
		\caption{The overall architecture of CapDet. The training paradigm of CapDet contains open-world object detection pre-training and dense captioning. In detection, CapDet contains a dual vision-language encoder. The image encoder generates region embeddings from detection and dense captioning data. The regression loss and centerness loss are introduced to regress the locations. The text encoder takes the category concepts as input to generate the embeddings from the [EOS] token. Then we treat the detection task as a matching task and adopt an alignment loss for the category embeddings and region embeddings. In dense captioning, an additional dense captioning head is proposed to take the region embeddings as input and generate the textual captions for corresponding regions with natural language.
		}
		\vspace{-4mm}
		\label{fig:framework}
\end{figure*}

\section{Method}
The overview of our proposed CapDet is shown in Figure \ref{fig:framework}. To construct a detector to either predict under a given category list or directly generate the concepts of predicted bounding boxes, we incorporate detection data and dense caption data together. In this section,  we will present a unified data format for the detection data and dense caption data in Section~\ref{sec:formulation}, the model architecture and pre-training objectives for open-world object detection pre-training in Section~\ref{sec:owd} and dense captioning in Section~\ref{sec:denseCap}.


\subsection{Unified Formulation}
\label{sec:formulation}
We defined a unified triplet-wise data format $(x, \{\textbf{b}_i\}^N_{i=1}, y^N_{i=1})$ for each sample from different sources. Specifically, $x \in \mathbb{R}^{3 \times h \times w}$ is the input image, $\{\mathbf{b}_i | \mathbf{b}_i\in\mathbb{R}^4\}_{i=1}^N$ denotes the bounding boxes coordinates for each region of the image, and the $y^N_{i=1}$ represents the concepts of the corresponding boxes.
$N$ denotes the number of regions.
A concept $y_i$ formatted as a sentence contains the category and textual description of the corresponding region. In detection data, a concept $y$ consists of the category name and the corresponding definition from the concept dictionary \cite{yao2022detclip}, while $y_i$ represents the region-grounded caption in dense caption data. 
For example, for an image $x$ in detection data, $y_i$ can be:
$$
y_i=``person,\ a\ human\ being."
$$
For an image $x$ from dense captioning data, $y_i$ can be: 
$$
y_i=``an\ outlet\ on\ the\ wall."
$$
With the triplet, we can learn a unified image-text alignment objective on the detection data and the dense captioning data. The unified formulation also ensures the joint training of open-world object detection pre-training and dense captioning. 


\subsection{Open-World Object Detection Pre-training}
\label{sec:owd}

Based on the unified formulation of detection data and dense captioning data, we regard the captions of regions in dense captioning data as a kind of category and utilize two different sources of data for the open-world object detection pre-training. Compared with the limited class list of detection data, dense caption data contains richer concepts and more semantic information than class names of individual regions. 
On the other hand, localization and recognition are two essential tasks of object detection. Traditional object detection always focuses on the salient objects in the image.
While the dense captioning data contains lots of annotations which are just parts of an object, \textit{e.g.}, \textit{an ear of an elephant}, it is not suitable to learn those annotations for the localization task.  Therefore, we only calculate the localization loss on detection data.

As shown in Figure \ref{fig:framework}, CapDet predicts the regions and treats the recognition task as a region-category matching task. 
For efficient learning on the matching task, we adopt the negative sampling proposed by~\cite{yao2022detclip} to provide negative concepts to enlarge the concept space in a batch.  Specifically, for each iteration, we randomly sample a negative concept set and add to the positive concept set (N samples) in a batch to obtain the final concept set $y^M_{i=1}$, where M represents the sum of the number of positive and negative samples. Finally, we format the triplet to $(x, \{\textbf{b}_i\}^N_{i=1}, y^M_{i=1})$.

CapDet contains a dual vision-language encoder and takes the triplet $(x, \{\textbf{b}_i\}^N_{i=1}, y^M_{i=1})$ as input. 
The image encoder $\Phi_v$  is an object detector that can predict the bounding boxes of regions from the input image $x$ and output the region features $O \in \mathbb{R}^{K \times D} $. The text encoder $\Phi_l$ takes the concept set $y^M_{i=1}$ as input and obtains the text embeddings $W \in \mathbb{R}^{M\times D}$ from the special token [EOS] concatenated with the text input.
$K, D$ denotes the number of predicted regions and region feature dimensions. The alignment score matrix $S \in \mathbb{R}^{K \times M}$ of regions and texts is calculated by:
\begin{equation}
O=\Phi_v(x), W=\Phi_l(y^M_{i=1}), S = OW^T 
\label{eqn:align}
\end{equation}
where $T $ denotes the transpose operation. A ground-truth alignment matrix $G \in \{0,1\}^{K\times M}$ is constructed to indicate the matching relation of regions and concepts. The alignment loss $\mathcal{L}_{align}$ is calculated by the predicted alignment scores of regions $S$ and the ground-truth alignment matrix $G$. Following \cite{li2022grounded,yao2022detclip}, we adopt the ATSS \cite{zhang2020bridging} detector as an image encoder, and $\mathcal{L}_{align}$ is typically a sigmoid focal loss. As a one-stage detector, the localization loss contains centeredness loss $\mathcal{L}_{cen}$ and bounding box regression loss $\mathcal{L}_{reg}$.
The training objective of detection pre-training can be written as:
\begin{equation}
\mathcal{L}= \begin{cases}\mathcal{L}_{\text {align }}+\alpha \mathcal{L}_{\text {reg }}+\beta \mathcal{L}_{\text {center }}, & \text { for detection } \\ \mathcal{L}_{\text {align }}, & \text { for dense captioning} 
\end{cases}
\end{equation}
where $\alpha$ and $\beta$ denote the weights for the centerness loss $\mathcal{L}_{cen}$ and box regression loss $\mathcal{L}_{reg}$, respectively. The $\mathcal{L}_{cen}$ is the sigmoid loss and the $\mathcal{L}_{reg}$ is the GIOU loss~\cite{rezatofighi2019generalized}.

\begin{table*}[h!]
\begin{centering} 
\par\end{centering}
\begin{center}
\begin{sc}
\begin{tabular}{c|cc|cccc}
\toprule 
\multirow{2}{*}{Model} & \multirow{2}{*}{Backbone} & \multirow{2}{*}{Pre-Train Data} & \multirow{2}{*}{Images Number} & \multicolumn{2}{c}{LVIS} \tabularnewline
 &  &  & & {AP} & {AP$_r$ / AP$_c$ / AP$_f$} \tabularnewline
  \midrule
  {Mask-RCNN~\cite{he2017mask}} & {Swin-T}  & {LVIS} & 0.1M & {34.1} & {19.1 / 34.0 / 37.0}  \tabularnewline
 {ATSS~\cite{zhang2020bridging}} & {Swin-T} & {LVIS} & 0.1M &  {33.6} & {19.7 / 32.4 / 37.2} \tabularnewline
 {ATSS~\cite{zhang2020bridging}} & {Swin-L} & {LVIS} & 0.1M & {43.9} & {30.6 / 43.7 / 46.3} \tabularnewline
 \midrule
 {MDETR~\cite{kamath2021mdetr}} & {RN101} & {GoldG+} & 0.77M & {24.2} & {20.9 / 24.3 / 24.2} \tabularnewline
 {GLIP-T(A)~\cite{li2022grounded}} & {Swin-T+DH+F} & {O365} & 0.66M & {18.5} & {14.2 / 13.9 / 23.4} \tabularnewline
 
  {GLIP-T(C)~\cite{li2022grounded}} & {Swin-T+DH+F} & {O365,GoldG} & 1.43M & {24.9} & {17.7 / 19.5 / 31.0} \tabularnewline
   
 {GLIP-T~\cite{li2022grounded}} & {Swin-T+DH+F} & {O365,GoldG,Cap4M} &5.43M & {26.0} & {20.8 / 21.4 / 31.0} \tabularnewline
{K-Lite~\cite{shen2022k}} & {Swin-T} & {O365} & 0.66M & {21.3} & {14.8 / 18.6 / 24.8} \tabularnewline
{K-Lite~\cite{shen2022k}} & {Swin-T} & {O365,GoldG} & 1.43M & {26.1} & {17.2 / 24.6 / 29.0} \tabularnewline
 
  {GLIPv2-T~\cite{zhang2022glipv2}} & {Swin-T+DH+F} & {O365,GoldG,Cap4M} & 5.43M & {29.0} & { \ \ \ - \ \ / \ \ \  - \ \ \ / \ \ \ -\ \ \ } \tabularnewline
 
  \midrule
{DetCLIP-T(A)~\cite{yao2022detclip}} & {Swin-T} & {O365} & 0.66M & {28.8} & {26.0 / 28.0 / 30.0}\tabularnewline
{DetCLIP-T(B)~\cite{yao2022detclip}} & {Swin-T} & {O365, GoldG} & 1.43M &{34.4} & {26.9 / 33.9 / 36.3}\tabularnewline
{DetCLIP-T(C)*~\cite{yao2022detclip}} & {Swin-T} & {O365, VG} & 0.73M &{31.5} & {27.5 / 30.6 / 33.0}\tabularnewline

\midrule
\rowcolor{mypink}{CapDet (Ours)} & {Swin-T} & {O365, VG} & \textbf{0.73M} & \textbf{{33.8}} & \textbf{{29.6 / 32.8 / 35.5}} \tabularnewline
\bottomrule

\end{tabular}
\caption{\label{tab:Preliminary-Experiment-1:}{Zero-shot performance on LVIS~\cite{gupta2019lvis} MiniVal5k datasets. AP$_r$ / AP$_c$ / AP$_f$ indicate the AP
values for rare, common, and frequent categories, respectively. “DH” and “F” in GLIP~\cite{li2022grounded} baselines stand for the dynamic head \cite{dai2021dynamic} and cross-modal fusion, respectively. Baselines with * are implemented with our code base.
GoldG+ denotes the GoldG plus the COCO~\cite{lin2014microsoft} caption dataset.
}}
\label{tab:ovdMain}
\end{sc}
\end{center}
\end{table*}
\vspace{-4mm}

\subsection{Dense Captioning}
\label{sec:denseCap}
The open-world object detection pre-training ensures CapDet gains the capacity to detect under given an arbitrary category list. However, when the given category list is not complete enough to cover the potential classes on a new domain data, the detector will perform worse on the categories which are not in the given list. Considering such limitation, we propose a dense captioning head $\Phi_C$ to generate semantically rich concepts with natural language for the predicted proposals.
In the dense captioning task, the model receives an image and produces a set of regions and the corresponding captions. 
The dense captioning head is a cross-modal decoder that takes the c predicted regions features $O$ generated by the image encoder as input.
The captioning (\textit{i.e.}, language modeling ) loss is calculated by:
\begin{equation}
\mathcal{L}_{cap}= -\log p(y_{it}|\Phi_c(y_{i(\tau<t)},O_i)),
\end{equation}
where $y_{it}$ means the t token in caption $y_i$ corresponding to region feature $O_i$, and $y_{i(\tau<t)}$ means tokens before t in caption $y_i$.
The overall pre-training loss can be written as:
\begin{equation}
\mathcal{L} = w_d\mathcal{L}_{det} +  w_c \mathcal{L}_{cap}, 
\end{equation}
where $w_d, w_c$ denote the weighting factor of $\mathcal{L}_{det}$ and $\mathcal{L}_{cap}$.

To minimize the gap in the type of bounding boxes between the detection data and dense captioning data, we propose a simple way to transform our detector as a class-agnostic detector and only select the top $k$ regions based on the centeredness scores to adapt to the dense captioning task. We can fine-tune our CapDet on the dense captioning data to achieve better performance.
Specifically, we propose "object" as the foreground concept and "background" as the background concept. The text encoder $\Phi_l$ outputs the concept embeddings $W' \in \mathbb{R}^{2\times D}$. Then the alignment scores $S' \in \mathbb{R}^{K \times 2}$ is calculated by Eqn.~\ref{eqn:align}. The captioning head takes the top $k$ most confident proposal embeddings based on centeredness scores as input to predict the region-grounded captions.






%% file: paper_files/4-experiment.tex
\begin{figure*}[t!]
		\begin{center}
            \includegraphics[ width=1.0\linewidth]{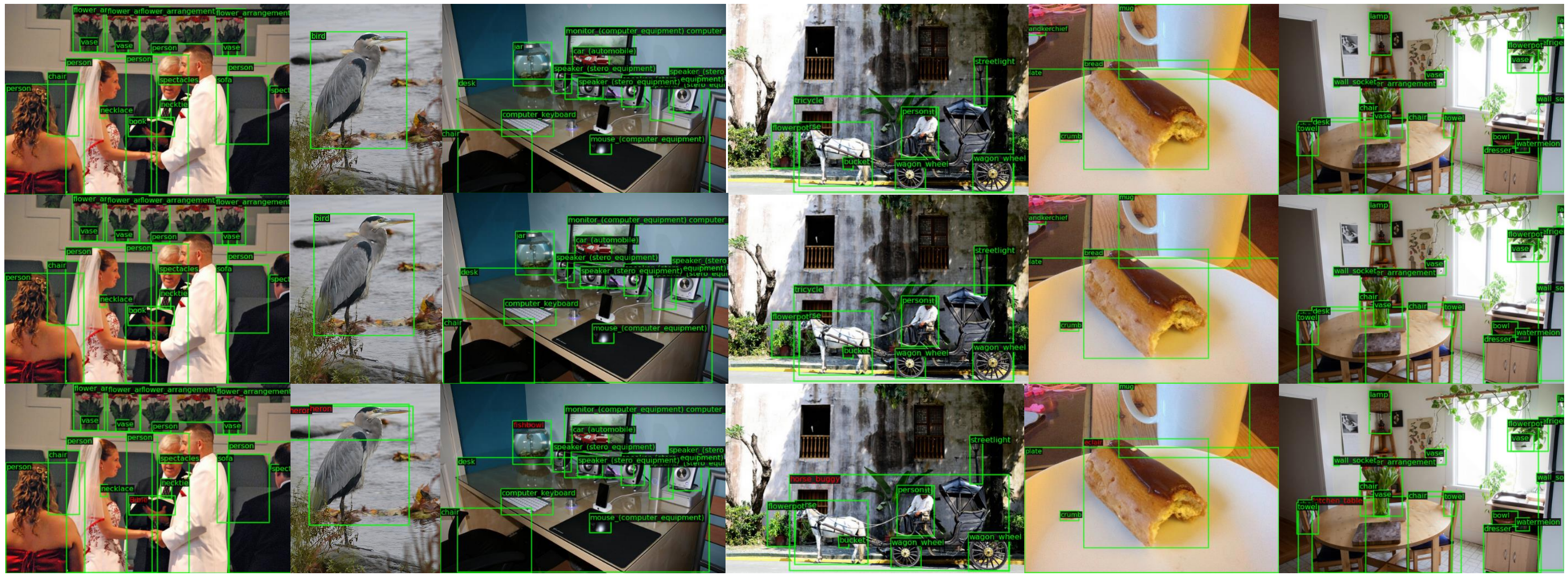}
		\end{center}
		\caption{Qualitative visualizations between GLIP-T, DetCLIP-T(C) and CapDet. From top to down, the three rows of images show the LVIS zero-shot detection results of GLIP-T, DetCLIP-T(C), and CapDet respectively. All models are pre-trained on O365 and VG.}
		\label{fig:novel detection}
\end{figure*}
\section{Experiment}
\paragraph{Implementation Details.}
For the image encoder, we adopt the Swin-T backbone proposed in Swin-Transformer~\cite{Liu2021SwinTH} which is pre-trained on ImageNet-1K~\cite{Deng2009ImageNetAL}. We use 12 layers 8 heads transformer as our text encoder and load a base model checkpoint released by FILIP \cite{Yao2022FILIPFI}, in order to make a fair comparison with DetCLIP~\cite{yao2022detclip}. The structure of the dense captioning head is consistent with that in the text encoder but trained from scratch for a fair comparison. We employ AdamW \cite{kingma2014adam} optimizer and set the batch size to 32. The learning rate is set to $1.4\times 10^{-4}$ for the parameters of the image encoder and detection head, and $1.4\times10^{-5}$ for the text encoder and dense captioning head. When fine-tuning the VG dataset to do the dense captioning task, we set the learning rate to $1.4\times10^{-4}$.  Without otherwise specified, all models are trained with 12 epochs and the learning rate is decayed with a factor of 0.1 at the 8-th and the 11-th epoch. The context token length for input text is set to 20. We set the number of input captions to 150, and the number of the region features N is determined by the feature map. The loss weight factor $w_c$ and $w_d$ are both set to 1.0. We build our model on MMDetection \cite{Chen2019MMDetectionOM} code base.
\paragraph{Dataset.}
Our CapDet is trained with two types of data, including detection data and caption data. Following DetCLIP \cite{yao2022detclip}, we use Object365 \cite{shao2019objects365} (it will be abbreviated as O365 in the following paper) as detection data, and sample 0.66M data from O365 v2 for training. Following GLIP \cite{li2022grounded} and DetCLIP \cite{yao2022detclip}, LVIS\cite{gupta2019lvis} MiniVal5k (defined in \cite{kamath2021mdetr})  which has 5000 images is used for detection evaluation. Moreover, we remove the training samples contained in the LVIS dataset for fair zero-transfer evaluation.
For dense captioning data, we mainly conduct our experiments on VG~\cite{krishna2017visual} V1.2 and VG-COCO (defined in ~\cite{shao2022region}). Following~\cite{shao2022region}, we allocate 77398 images for training and 5000 images for validation and testing on VG. As demonstrated in~\cite{krishna2017visual}, the ground-truth bounding boxes of VG are much denser than the other object detection datasets, \textit{i.e.}, the average number of per sample in MS COCO~\cite{lin2014microsoft} is only 7.1 \textit{vs.} 35.4 in VG. Then an intersection of VG V1.2 and MS COCO is proposed by~\cite{shao2022region} and is denoted as VG-COCO, which has 38080 images for training, 2489 for validation, and 2476 for testing. 

\paragraph{Benchmark Settings.}
We mainly evaluate our method on open-vocabulary object detection and dense captioning task. For open-vocabulary object detection, we evaluate the direct domain transfer on LVIS~\cite{gupta2019lvis} which contains 1203 categories. Following ~\cite{yao2022detclip, li2022grounded}, we metric the zero-shot detection performance by the Average Precision (AP) on a 5k subset.
The annotations of LVIS data are split into three folds, \textit{i.e.}, rare, common, and frequency, based on the number of categories. Since there is almost no overlap between the rare classes and the classes of training dataset Objects365~\cite{shao2019objects365}, the AP of the rare classes shows a valuable zero-shot detection performance. 
For the dense captioning task, we follow the setting of~\cite{johnson2016densecap} to evaluate the VG and VG-COCO. The evaluation metric we adopt is the mean Average Precision proposed by~\cite{johnson2016densecap} which is calculated across a range of thresholds for both localization and language accuracy, \textit{i.e.}, the intersection over union (IOU) thresholds .3, .4, .5, .6, .7 are used for localization and the METEOR score' thresholds 0, .05, .1, .15, .2, .25 is adopted for evaluating the language generation. 
\begin{figure*}[t!]
		\begin{center}
            \includegraphics[width=0.99\linewidth]{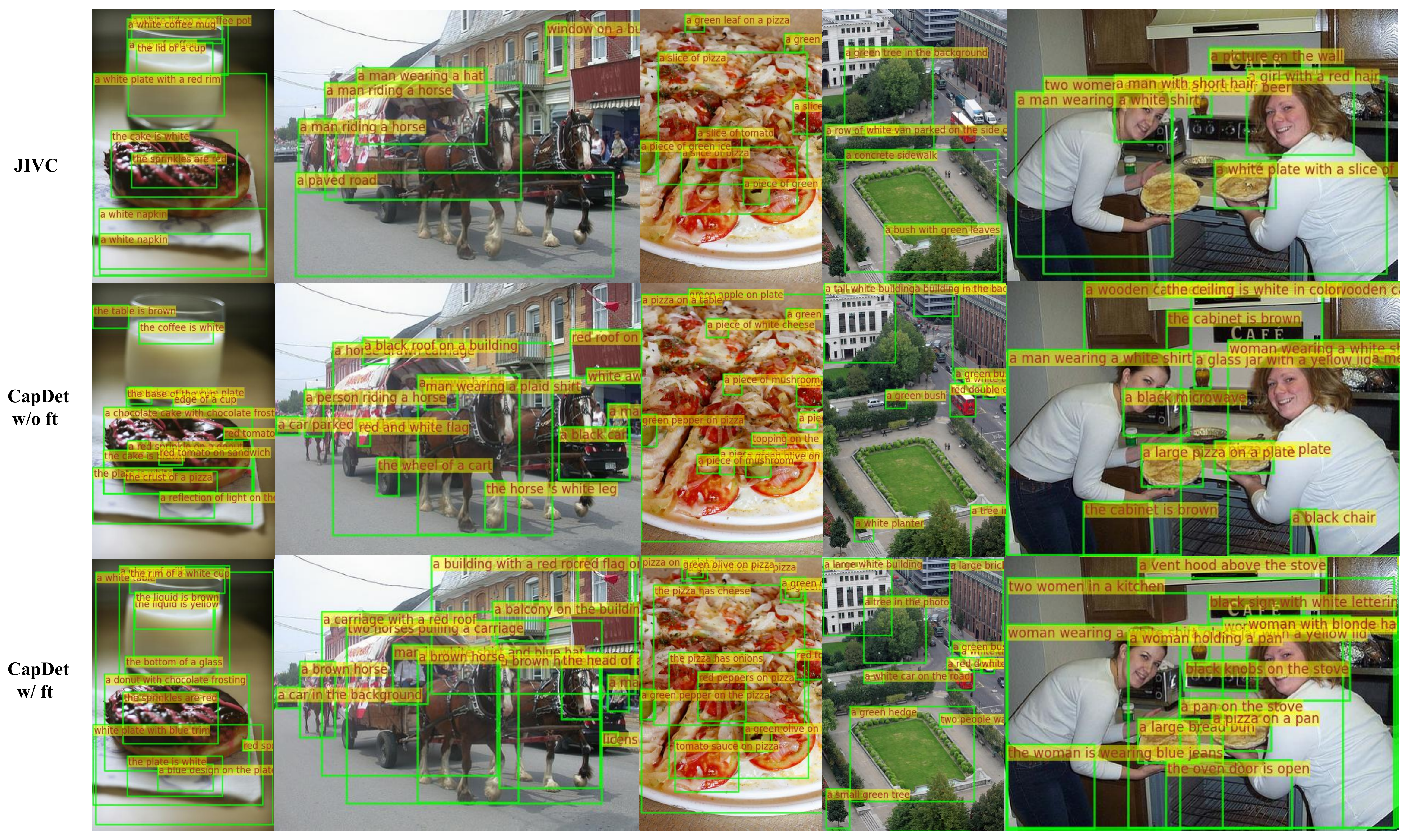}
		\end{center}
		\vspace{-3mm}
		\caption{Qualitative visualizations between JIVC and CapDet. “w/o ft” means do caption without finetune, while “w/ ft” means with finetune.}
		\label{fig:vgcaption result}
\vspace{-6mm}
\end{figure*}

\vspace{-1mm}
\subsection{Open-world Detection Results}
Table~\ref{tab:ovdMain} shows our zero-shot object detection performance on LVIS. We mainly train our CapDet with the backbone Swin-T~\cite{Liu2021SwinTH} on the detection data Objects365~\cite{shao2019objects365} and dense captioning data (VG~\cite{krishna2017visual}). 
Since DetCLIP does not report the performance on O365 and VG, we train DetCLIP on the two datasets under the same settings and denote it as DetCLIP-T(C) for a fair comparison. 
Comparing the 11th row and 12th row, our CapDet outperforms DetCLIP-T(C) on the same data scale and backbone with an extra simple caption head. 
Moreover, our model's zero-shot performance even surpasses the fully-supervised model with the same backbone by a large margin on rare classes, \textit{i.e.}, CapDet outperforms ATSS by 9.9\%. 
\vspace{-5mm}

\paragraph{Qualitative Visualizations}
Figure.~\ref{fig:novel detection} illustrates the detection results on LVIS~\cite{krishna2017visual} dataset from GLIP-T, DetCLIP-T(C), and CapDet. All three models are trained on O365 and VG, and details are given in Section 4.3. Given a category list, the rare classes are detected more precisely by our CapDet, \textit{e.g.}, “kitchen table” in the first column, “horse buggy” in the third column, and “fishbowl” in the sixth column that our model CapDet detects correctly but the other two not.

\subsection{Dense Captioning Results}
Due to the target bounding boxes in dense captioning data containing lots of local structures of objects and being much denser than the bounding boxes in object detection data, we do not regress the bounding box in the pre-training stage. The previous works directly train on the dense captioning data and generate captions on the top $k$ proposals ranking by a confidence score. When fine-tuning our model on the VG dataset for the dense captioning tasks, we transform our CapDet into a class-agnostic detector. Specifically, we propose “object" as the foreground concept and “background" as the background concept for computing alignment scores. The scores are used as proposal confidences to predict the region-grounded captions.

Table~\ref{tab2:SOTA on denseCap VG} and Table~\ref{tab2:SOTA on denseCap VG-COCO} show CapDet significantly outperforms the latest work TDC~\cite{shao2022region}  by 2.5\% on mAP on VG and TDC+ROCSU~\cite{shao2022region} by 2.08\%, respectively. It is worth noticing that, even against given the ground-truth bounding boxes with the previous method COCG~\cite{li2019learning} denoted as COCOG\&GT, our CapDet still gains a 43.80\% mAP increase and achieves state-of-the-art. One important reason is that the excellent detection performance of our model assists the localization ability of dense captioning tasks.

\begin{table}[t!]
\begin{center}
\begin{tabular}{l|c}
\toprule 
{Method} & mAP(\%) \tabularnewline
\midrule
FCLN~\cite{johnson2016densecap} &5.16 \tabularnewline
{JIVC~\cite{yang2017dense}}& 9.96\tabularnewline
ImgG~\cite{li2019learning} & 9.68\tabularnewline
COCD~\cite{li2019learning} & 9.75 \tabularnewline
{COCG~\cite{li2019learning}} & 10.39\tabularnewline
{CAG-Net~\cite{yin2019context}}& 10.51\tabularnewline
{TDC~\cite{shao2022region}} &11.90\tabularnewline
\midrule
\rowcolor{mypink}{CapDet (Ours)}& \textbf{15.44} \tabularnewline
\bottomrule
\end{tabular}
\caption{\label{tab2:SOTA on denseCap VG}{Comparison of mAP (\%) performance on dense captioning benchmark on the VG V1.2 dataset.}}
\vspace{-2em}
\end{center}
\end{table}

\begin{table}[t!]
\begin{center}
\begin{tabular}{l|c}
\toprule 
{Method} & mAP(\%) \tabularnewline
\midrule
FCLN~\cite{johnson2016densecap} &4.23 \tabularnewline
{JIVC~\cite{yang2017dense}}& 7.85\tabularnewline
Max Pooling~\cite{li2019learning} & 7.86 \tabularnewline
ImgG~\cite{li2019learning} & 7.81\tabularnewline
COCD~\cite{li2019learning} & 7.92 \tabularnewline
{COCG~\cite{li2019learning}} & 8.90\tabularnewline
COCG-LocSiz~\cite{li2019learning} & 8.76 \tabularnewline
COCG\&GT~\cite{li2019learning} & 9.79 \tabularnewline
{TDC+ROCSU~\cite{shao2022region}} &11.9\tabularnewline
\midrule
\rowcolor{mypink}{CapDet (Ours)}& \textbf{13.98} \tabularnewline
\bottomrule
\end{tabular}
\caption{\label{tab2:SOTA on denseCap VG-COCO}{Comparison of mAP (\%) performance on the dense captioning benchmark on the VG-COCO Dataset.}}\vspace{-3mm}
\end{center}
\end{table}

\noindent\textbf{Qualitative Visualizations.}
Figure \ref{fig:vgcaption result} shows a qualitative visualization comparison between JIVC \cite{yang2017dense} and our CapDet. The three image rows from top to bottom are the visualization of JIVC, CapDet without fine-tuning, and CapDet with finetuning. In the second row, CapDet can locate more objects than JIVC, owing to our model's superior localization performance. After finetuning, CapDet can further describe a region rather than a single object such as “two women in a kitchen" in the 5-th column.

\subsection{Ablation Studies}
\subsubsection{Ablations for Unified Pre-training}
\textbf{Effect on different baselines.}
Table~\ref{tab3:Effect on different baseline} investigates the advantages of dense captioning heads on different baselines. We integrate our dense captioning head with GLIP-T or DetCLIP-T. The GLIP-T is implemented with parallel text encoding without external knowledge following the setting as ablations in~\cite{shen2022k} on our code base. All the results are pre-trained on Objects365 and VG. The results show that our dense captioning head is able to boost the generalization and model-agnostic. 

\begin{table}[t!]
\begin{centering} 
\par\end{centering}
\begin{center}
\begin{sc}
\resizebox{\linewidth}{!}{
\begin{tabular}{c|c|c|cccc}
\toprule 
\multirow{2}{*}{Model}  & \multirow{2}{*}{DC Head} & \multicolumn{2}{c}{LVIS} \tabularnewline

   &  & {AP} & {AP$_r$ / AP$_c$ / AP$_f$}  \tabularnewline
  \midrule
 {GLIP-T} & {\xmark}  & {30.4} & {22.5 / 29.0 / 33.0} \tabularnewline
 
  {GLIP-T} & {\cmark} &{33.1}& {27.0 / 32.1 / 35.0} \tabularnewline
   
 {DetCLIP-T} & {\xmark} & {31.5} & {27.5 / 30.6 / 33.0}& \tabularnewline
 {DetCLIP-T} & {\cmark} & {33.8} & {29.6 / 32.8 / 35.5}& \tabularnewline
 
\bottomrule
\end{tabular}
}
\caption{\label{tab3:Effect on different baseline}{Ablations on integrating our dense captioning head into different baselines.}}
\end{sc}
\end{center}
\vspace{-2em}
\end{table}

\begin{table}[t!]
\begin{center}
\resizebox{\linewidth}{!}{
\begin{tabular}{c|c|c|c}
\toprule 
{Pre-training Data} & Fine-tune & DCap mAP(\%)& Box mAP(\%)\tabularnewline
\midrule
{VG}& \xmark & 12.86 & 27.65\tabularnewline
{O365,VG}& \xmark & 4.72 & 9.65\tabularnewline
VG & \cmark &  13.83 & 28.58 \tabularnewline
{O365,VG} & \cmark & 15.44 & 30.61\tabularnewline
\bottomrule
\end{tabular}
}
\caption{\label{tab5:Ablations on denseCap task}{Ablations on incorporating data from different sources. “DCap” stands for the dense caption mAP.}}\vspace{-3mm}
\end{center}
\end{table}

\noindent\textbf{Effect of dense captioning data.}
Table~\ref{tab:ovdMain} shows the efficiency of incorporating dense captioning data. Specifically, only 0.07M data added, the DetCLIP-T(C) gains +2.7\% overall AP and +1.5\% $AP_r$ on LVIS compared to DetCLIP-T(A). The performance of DetCLP-T(A) on rare categories also outperforms DetCLIP-T(C) train on Objects365 and GOLDG, while the data size is 1.43M \textit{vs.} 0.73M.


 
   
 
 

\vspace{-3mm}
\subsubsection{Ablations for dense captioning}

We investigate the impact of training policy and data from different sources on the dense captioning task.
As shown in row1 in  Table~\ref{tab5:Ablations on denseCap task}, our CapDet still achieves a significant performance which is directly trained on VG outperforms the previous task (\textit{i.e.}, TDC~\cite{shao2022region} in Table \ref{tab2:SOTA on denseCap VG}). Row2 is our CapDet and is pre-trained on Objects365 and VG, while only the bounding box in the Objects365 is regressed, and then transformed on a dense captioning task.  Since the type of bounding boxes in dense captioning is different from the detection data, the result of the direct transforming to dense captioning is worse. However, we've proved that our model still keeps the dense captioning capacity on the salient objects in Figure~\ref{fig:vgcaption result}. The results in  row3 and row4 indicate that pre-training on the detection data Objects365 is also beneficial to the dense captioning task.

%% file: paper_files/5-conclusion.tex

\vspace{-3mm}
\section{Limitations}
These are a few issues that we need to improve in the future: (1) Although our unification training paradigm works well on open-vocabulary object detection and dense captioning task, the training of dense captioning generation costs lots of time. 
(2)
In addition, existing dense captioning data is high-cost to collect. We will research how to collect large-scale dense captioning data by auto annotation and get better performance with the scaled-up data.

\vspace{-2mm}
\section{Conclusion}
 In this paper, we propose a novel open-world object detection method named CapDet. Our CapDet is more practical in the open world and real scenes. Specifically, CapDet introduces a unification training framework including open-world object detection pre-training and dense captioning. The unification enables our CapDet to localize and recognize concepts in an arbitrary given category list or directly generate textual captions for predicted new concept objects. Experiments show that the design of unification is both beneficial to open-world object detection tasks and dense captioning tasks. 
In the future, our CapDet can be easily injected into any open world and real scenes tasks. The unification framework can also be integrated into any other OWD/OVD methods to generate semantic-rich concepts for unknown/novel objects.

\vspace{-2mm}

\renewcommand{\thefootnote}{\arabic{footnote}}
\paragraph{Acknowledgements} We gratefully acknowledge the support of MindSpore\footnote{https://www.mindspore.cn/}, CANN (Compute
Architecture for Neural Networks) and Ascend AI Processor used for this research.

%% file: paper_files/appendix.tex


\clearpage

\onecolumn
\section*{Appendix for CapDet: Unifying Dense Captioning and Open-World Detection Pretraining}
\maketitle

\appendix

\section{Detailed Experimental Settings}
The detailed architecture parameters for different modules of CapDet are shown in Table~\ref{tab1:model params}.
For the learning rate scheduler, we assign a base learning rate and then linearly warm it up to the peak learning rate according to the effective total batch size by a square root strategy, $lr_{peak}$ = $lr_{base}\times \sqrt{ \rm batchsize/ 16}$, \textit{e.g.}, we set image encoder base learning rate to $1\times10^{-4}$ and it automatically scales to $1.4\times10^{-4}$. The training hyperparameters used for CapDet are shown in Table~\ref{tab2:hyperparameter}.
\vspace{5mm}

\begin{minipage}{\textwidth}

\begin{minipage}[t]{0.48\textwidth}
\makeatletter\def\@captype{table}
\begin{centering}

\begin{tabular}{l|c}
\toprule 
{Image Encoder} & Value \tabularnewline
\midrule
backbone & swin-t \tabularnewline
neck & fpn\tabularnewline
input resolution &1333$\times$800 \tabularnewline
\midrule
{Text Encoder} & Value \tabularnewline
\midrule
width & 512 \tabularnewline
heads & 8\tabularnewline
layers & 12\tabularnewline
\midrule
{Cross-Modal Decoder} & Value \tabularnewline
\midrule
width & 512 \tabularnewline
heads & 12\tabularnewline
layers & 12\tabularnewline
\bottomrule
\end{tabular}
\caption{\label{tab1:model params}{Detailed architecture parameters for different module.}}\vspace{-3mm}
\end{centering}
\end{minipage}
\begin{minipage}[t]{0.48\textwidth}
\makeatletter\def\@captype{table}
\begin{centering}

\begin{tabular}{l|c}
\toprule 
{Hyperparameter} & Value(\%) \tabularnewline
\midrule
Image encoder lr & $1.4\times10^{-4}$ \tabularnewline
Text encoder lr & $1.4\times10^{-5}$ \tabularnewline
Crossmodal decoder lr & $1.4\times10^{-5}$ \tabularnewline
Learning policy & CosineAnnealing \tabularnewline
warmup ratio & 0.0001 \tabularnewline
warmup iters & 1000 \tabularnewline
batchsize & 32 \tabularnewline
weight decay & 0.05 \tabularnewline
$w_c$ & 1 \tabularnewline
$w_d$ & 1 \tabularnewline
\bottomrule
\end{tabular}
\vspace{5mm}
\caption{\label{tab2:hyperparameter}{The training hyperparameters used for CapDet.}}\vspace{-3mm}
\end{centering}
\end{minipage}
\end{minipage}


\vspace{2mm}

\section{Fine-tuning Results on LVIS}
We provide the fine-tuning results on LVIS in Table~\ref{tab:R3Q2} below. 
We observe that CapDet outperforms the baseline DetCLIP with 1.2\% AP on average and 6.5\% AP on rare classes.
Besides, though pre-trained with fewer data and tasks, CapDet shows a competitive performance compared with the GLIPv2.
\begin{table}[th]
\begin{centering} 
\par\end{centering}
\begin{center}
\begin{sc}
\begin{tabular}{c|cc|cccc}
\toprule 

\multirow{2}{*}{Model} & \multirow{2}{*}{Backbone} & \multirow{2}{*}{Pre-Train Data} & \multirow{2}{*}{Images Number} & \multicolumn{2}{c}{LVIS} \tabularnewline
  &  &  & & {AP} & {AP$_r$ / AP$_c$ / AP$_f$} \tabularnewline
  \midrule

{DetCLIP-T(C)*}~\cite{yao2022detclip} & Swin-T & O365, VG & 0.73M &45.6
&	{33.6 / 45.8 / 47.5} \tabularnewline
GLIPv2-T~\cite{li2022grounded} & Swin-T+DH+F & O365, GoldG, Cap4M  & 5.43M & 50.6  &  \ \ \ - \ \ / \ \ \  - \ \ \ / \ \ \ -\ \ \ \tabularnewline

\rowcolor{mypink}CapDet (Ours) & Swin-T & \textbf{O365, VG }& \textbf{0.73M }& \textbf{47.2}
 &	{\textbf{40.1} / \textbf{46.9} / \textbf{48.7}} \tabularnewline
\bottomrule

\end{tabular}
\caption{\label{tab:R3Q2}{Fine-tuning performance on LVIS~\cite{gupta2019lvis} MiniVal5k datasets. AP$_r$/AP$_c$/AP$_f$ indicate the AP
values for rare, common, and frequent categories. `DH' and `F' in GLIP~\cite{li2022grounded} baselines stand for the dynamic head \cite{dai2021dynamic} and cross-modal fusion.  }}
\end{sc}
\end{center}
\vspace{-4mm}
\end{table}

\section{Open-World Detection Results on LVIS Full Validation Set}
Table~\ref{tab:LVIS_full} reports our zero-shot object detection performance on LVIS~\cite{gupta2019lvis} full validation set.
Following~\cite{yao2022detclip,li2022grounded}, we take the class names with additional manually designed prompts as input of text encoder. Comparing the 5th row and 6th row, our CapDet still outperforms DetCLIP-T(C) on the same data scale and backbone with an extra simple caption head. The zero-shot performance surpasses the previous methods with the same backbone by a large margin on rare classes, \textit{e.g.}, CapDet trained on fewer data outperforms GLIP-T~\cite{li2022grounded} by 10.8\% on AP$_r$.
\begin{table}[h!]
\begin{centering} 
\par\end{centering}
\begin{center}
\begin{small}
\begin{sc}
\resizebox{0.9\textwidth}{!}{
\begin{tabular}{c|ccc|cccc}
\toprule 
\multirow{2}{*}{Model} & \multirow{2}{*}{Backbone} & \multirow{2}{*}{Pre-Train Data} &\multirow{2}{*}{Images Number}& \multicolumn{2}{c}{LVIS Val Full}  \tabularnewline
 &  &  & & {AP} & {AP$_r$ / AP$_c$ / AP$_f$}  \tabularnewline
  \midrule

 {GLIP-T(A)\cite{li2022grounded}} & {Swin-T+DH+F} & {O365} & 0.66M & {12.3} & {6.00 / 8.00 / 19.4}  \tabularnewline
 {GLIP-T\cite{li2022grounded}} & {Swin-T+DH+F} & {O365,GoldG,Cap4M} & 5.43M & {17.2}& {10.1 / 12.5 / 25.2} \tabularnewline
  \midrule
{DetCLIP-T(A)~\cite{yao2022detclip}} & {Swin-T} & {O365} & 0.66M & {22.1} & {18.4 / 20.1 / 26.0} \tabularnewline
{DetCLIP-T(C)~\cite{yao2022detclip}} & {Swin-T} & {O365, VG} & 0.73M & {23.5} &{18.4 / 21.6 / 27.9}   \tabularnewline
\midrule
\rowcolor{mypink} {CapDet} (Ours) & {Swin-T} & {O365, VG} & 0.73M & {\textbf{26.1}} &{\textbf{20.9} / \textbf{24.4} / \textbf{30.2}}\tabularnewline
\bottomrule
\end{tabular}
}
\caption{{Zero-shot transfer performance on LVIS~\cite{gupta2019lvis} full validation dataset. AP$_r$/AP$_c$/AP$_f$ indicates the AP
values for rare, common, and frequent categories. `DH' and `F' in GLIP~\cite{li2022grounded} baselines stand for the dynamic head \cite{dai2021dynamic} and cross-modal fusion. 
}}
\label{tab:LVIS_full}
\end{sc}
\end{small}
\end{center}
\end{table}

\section{Analysis of the Improvements on OVD}
We attribute the improvements on OVD to the reason that the incorporation of captioning head brings more generalizability for the region features, which in turn helps the learning of OVD task.
Specifically, the dense captioning task is essentially a sequential classification task with a large enough class space (\textit{i.e.}, word tokens), while alignment task is a single-step classification task with a limited class space. Therefore, training with dense captioning tasks will bring the region feature into a more proper location in feature space rather than simply pulling them together via only detection task.
As shown in Table~\ref{tab:R2Q2}, we further conduct the experiments to demonstrate the effectiveness of pre-training under captioning.
By comparing the row 2 and 5,  we observe that even with only dense captioning data~(VG data), pre-training with the dense captioning paradigm also brings a significant improvement.
\vspace{-2pt}
\begin{table}[th]
\begin{centering} 
\par\end{centering}
\begin{center}
\begin{sc}
\begin{tabular}{c|c|cccc}
\toprule 

\multirow{2}{*}{Model} & \multirow{2}{*}{Pre-Train Data} & \multicolumn{2}{c}{LVIS}  \tabularnewline
 &  & {AP} & {AP$_r$ / AP$_c$ / AP$_f$}  \tabularnewline
  \midrule
  
\multirow{3}{*}{DetCLIP-T~\cite{yao2022detclip}} & {O365} & 28.8 &{26.0 / 28.0 / 30.0} & \tabularnewline
& VG & 10.3 & {8.6 / 10.1 / 10.8}& \tabularnewline
& {O365, VG} &{31.5} & {27.5 / 30.6 / 33.0}\tabularnewline

\midrule
\multirow{3}{*}{CapDet} & {O365} & 28.5	
&{25.2 / 27.5 / 29.9} \tabularnewline
& VG       & 11.4 & {10.2 / 11.1 / 11.8} & 	
\tabularnewline
& O365, VG &{\textbf{33.8}} & {\textbf{29.6} / \textbf{32.8} / \textbf{35.5}}\tabularnewline
\bottomrule

\end{tabular}
\caption{\label{tab:R2Q2}{Zero-shot performance on LVIS~\cite{gupta2019lvis} MiniVal5k datasets. AP$_r$ / AP$_c$ / AP$_f$ indicate the AP
values for rare, common, and frequent categories, respectively. “DH” and “F” in GLIP~\cite{li2022grounded} baselines stand for the dynamic head \cite{dai2021dynamic} and cross-modal fusion, respectively.}}
\end{sc}
\end{center}
\vspace{-4mm}
\end{table}
\vspace{-7pt}

\section{`Real' Open-world Object Detection Deployment Strategy}
\begin{figure*}
		\begin{center}
            \includegraphics[width=0.7\linewidth]{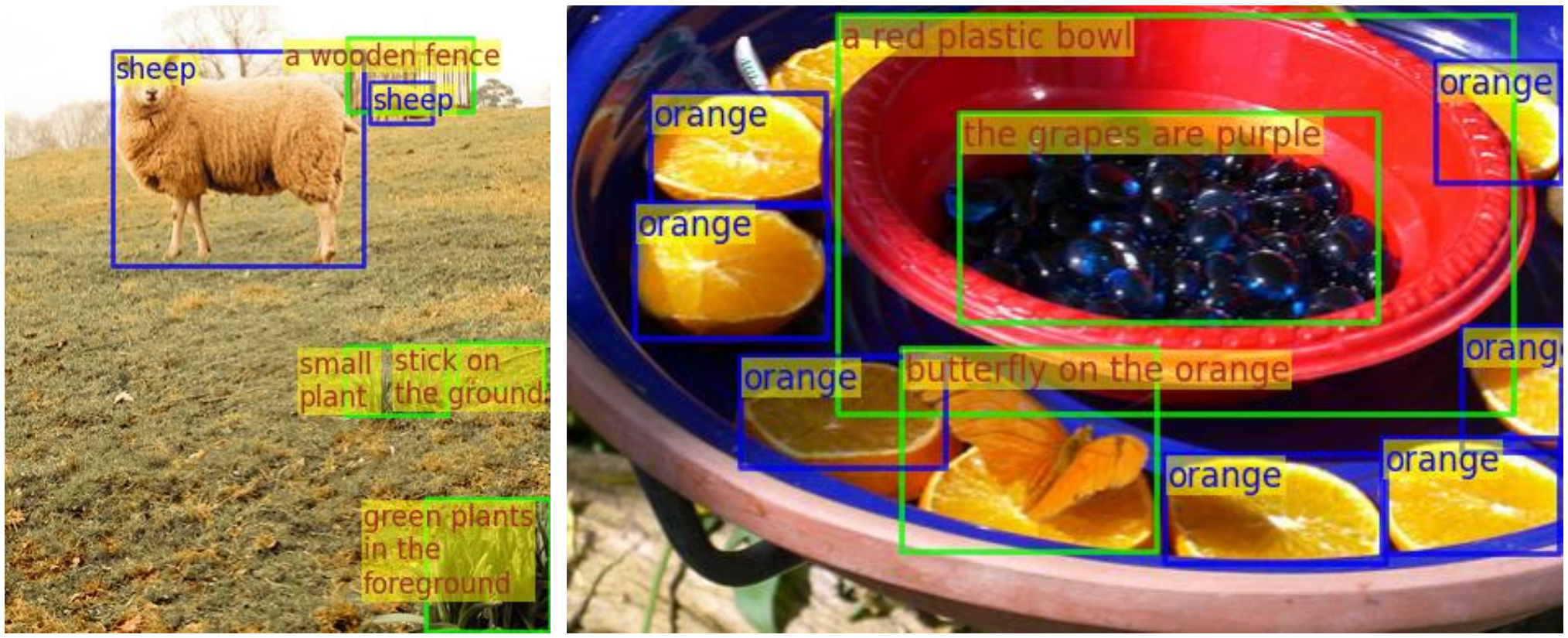}
		\end{center}
		\vspace{-3mm}
		\caption{Deployment results.}
		\label{fig:deployment vis}
\vspace{-4mm}
\end{figure*}

In this paper, the detection and dense captioning task are illustrated separately for better understanding and comparison with other methods, since no benchmark has considered combining these two tasks.
For the practical deployment, we propose a simple two-stage ensemble way to stay true to the motivation.
Specifically, in the first stage, we execute detection on images among the pre-defined categories list and treat the proposals with maximum alignment scores among all classes less than a threshold as `unknown' objects. 
Then in the second stage, we generate the captions for the `unknown' objects. 
To demonstrate the effectiveness of the proposed strategies, We conduct detection on the images with 80 categories of COCO and regenerate captions for the `unknown' objects. As shown in the Figure~\ref{fig:deployment vis} , our proposed strategy expands the semantic space of the limited categories list and shows reasonable results.

\section{More Ablation Studies}





\paragraph{Ablations on Pre-trained Language Model}
Table~\ref{tab:languageModel} reports the effect of different tokenizers and pre-trained language models loaded for text encoder. We ablate two kinds of pre-trained language models and corresponding tokenizers for our text encoder. For dense captioning head, we construct the same decoder as BLIP~\cite{li2022blip} decoder and keep the tokenizer the same as the text encoder. The results indicate the FILIP~\cite{Yao2022FILIPFI} encoder with byte pair encoding performs a better generalization, since it is pre-trained on a larger scale of data, \textit{i.e.}, 300M in FILIP~\cite{Yao2022FILIPFI} \textit{vs.} 128M in BLIP~\cite{li2022blip}.

\begin{table}[th]
\begin{centering}

\begin{tabular}{ccc|c|c|cccc}
\toprule 
\multirow{2}{*}{Pre-trained Model}&
\multirow{2}{*}{Tokenizer} & 
\multirow{2}{*}{Vocab Size}    & \multirow{2}{*}{DC Head}&\multicolumn{2}{c}{LVIS} \tabularnewline

 &  &  & & {AP} & {AP$_r$ / AP$_c$ / AP$_f$}  \tabularnewline
  \midrule
\multirow{2}{*}{BLIP~\cite{li2022blip}} &\multirow{2}{*}{WordPiece} &  \multirow{2}{*}{30524} & {\xmark} & {30.4} & {26.7 / 29.4 / 32.0}& \tabularnewline
 & &  & {\cmark} & {32.4} & {27.4 / 31.8 / 33.9}& \tabularnewline
 \midrule
\multirow{2}{*}{FILIP~\cite{Yao2022FILIPFI}} & \multirow{2}{*}{BPE} &  \multirow{2}{*}{49408} & {\xmark}  & {31.5} & {27.5 / 30.6 / 33.0} \tabularnewline
 
 &  & & {\cmark} &{33.8}& {29.6 / 32.8 / 35.5} \tabularnewline
\bottomrule
\end{tabular}

\caption{\label{tab:languageModel}{Effect of different tokenizers and language models. `DC Head' and `BPE' stand for the integration of Dense Captioning Head and Byte Pair Encoding.}}
\end{centering}
\end{table}

 

\paragraph{Ablations on the Weighting Factor of Dense Captioning Loss}
We study the effect of weights of detection loss and dense captioning loss during pre-training. We set the weighting factor of detection loss $w_d$ to 1.0. Table~\ref{tab:lossweight} provides the ablations of the weighting factor of dense captioning loss $w_c$. We choose $w_c = 1$ for CapDet, since the result of overall AP is the best.
\begin{table}[h!]

\begin{centering}

\begin{tabular}{c|ccc}
\toprule 
\multirow{2}{*}{$w_c$} &  \multicolumn{2}{c}{LVIS} \tabularnewline
 & {AP} & {AP$_r$ / AP$_c$ / AP$_f$} \tabularnewline
  \midrule
  0.5 &  {33.6} & {31.0 / 32.8 / 34.9}  \tabularnewline
 1.0 &  \textbf{{33.8}} & {29.6 / 32.8 / 35.5} \tabularnewline
 1.5 & {33.5} & {32.0 / 32.1 / 35.0} \tabularnewline
 \midrule
\end{tabular}
\caption{\label{tab:lossweight}{Effect of weighting factor of dense captioning loss.}}\vspace{-3mm}
\end{centering}
\end{table}


\section{More Qualitative Results}

\begin{figure*}[t!]
		\begin{center}
            \includegraphics[width=0.99\linewidth]{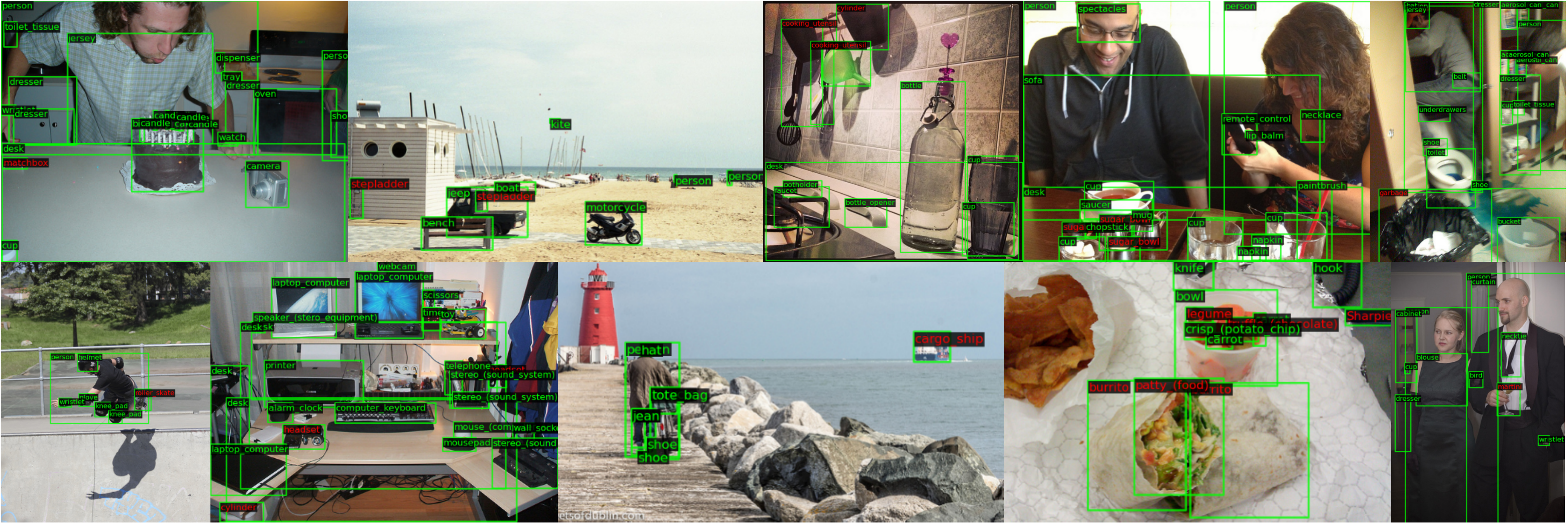}
		\end{center}
		\vspace{-3mm}
		\caption{Qualitative visualizations on LVIS.}
		\label{fig:detection vis}
\vspace{-4mm}
\end{figure*}

\paragraph{Open-World Detection Results}
Figure~\ref{fig:detection vis} illustrates more detection results on LVIS~\cite{gupta2019lvis} dataset from our CapDet. We highlight the detected rare classes's text in red.

\begin{figure*}[t!]
		\begin{center}
            \includegraphics[width=0.99\linewidth]{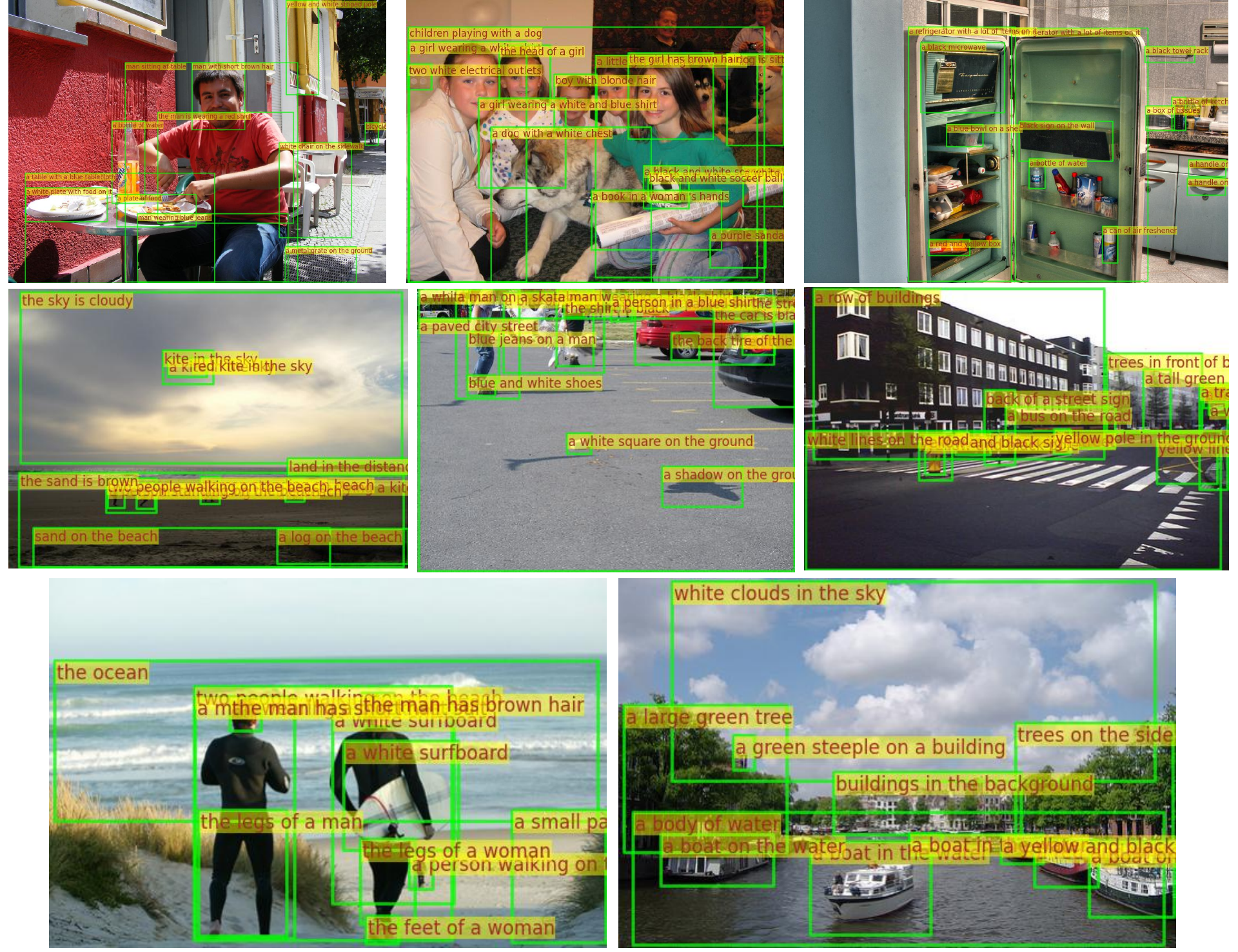}
		\end{center}
		\vspace{-4mm}
		\caption{Qualitative visualizations on VG.}
		\label{fig:vgcaption vis}
\vspace{-3mm}
\end{figure*}

\begin{figure*}[t!]
		\begin{center}
            \includegraphics[width=0.99\linewidth]{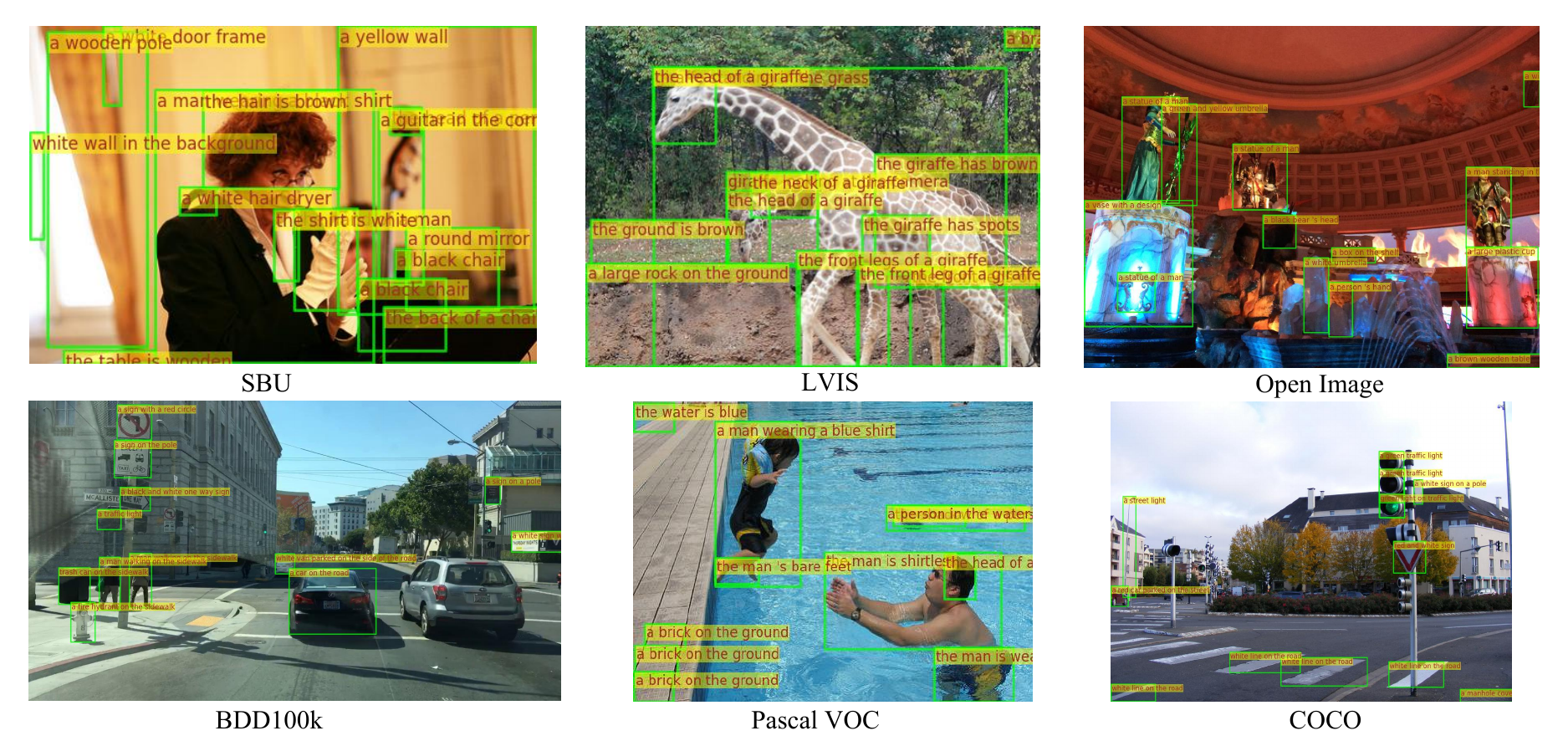}
		\end{center}
		\vspace{-4mm}
		\caption{Qualitative visualizations on several datasets.}
		\label{fig:openfig}
\vspace{-4mm}
\end{figure*}

\paragraph{Dense Captioning Results}
Figure \ref{fig:vgcaption vis} shows more captioning results on VisualGenome~\cite{krishna2017visual} dataset. Our model CapDet locates not only ``object" such as ``bicycle" but also ``region" such as ``a shadow on the ground". We also explored the zero-shot generalization ability of CapDet. We directly use our model to do the zero-shot dense captioning task without finetuning on serveral datasets, which include 
SBU~\cite{ordonez2011im2text}, LVIS~\cite{gupta2019lvis}, Open Image~\cite{kuznetsova2020open}, BDD100K~\cite{yu2020bdd100k}, Pascal VOC~\cite{everingham2015pascal} and COCO~\cite{chen2015microsoft}. As shown in Figure \ref{fig:openfig}, CapDet can accurately locate objects and generate corresponding region-grounded captions.

